\documentclass{article}

    \usepackage[preprint]{neurips_2025}

\usepackage[utf8]{inputenc} 
\usepackage[T1]{fontenc}    
\usepackage{hyperref}       
\usepackage{url}            
\usepackage{booktabs}       
\usepackage{amsfonts}       
\usepackage{nicefrac}       
\usepackage{microtype}      
\usepackage{amsmath}
\usepackage{booktabs}
\usepackage{graphicx}
\usepackage{multirow}

\title{VELVET-Med: Vision and Efficient Language Pre-training for Volumetric Imaging Tasks in Medicine}

\author{%
  Ziyang Zhang $^\dagger$\\
  MedVisAI Lab\\
  Department of ECE\\
  Northwestern University\\
  \And
  Yang Yu $^\dagger$\\
  Institute for Infocomm Research (I$^{2}$R)\\ 
  A*STAR, Singapore \\
  \AND
  Xulei Yang $^*$\\
  Institute for Infocomm Research (I$^{2}$R)\\ 
  A*STAR, Singapore\\
  \AND
  Si Yong Yeo $^*$\\
  MedVisAI Lab\\
  Lee Kong Chian School of Medicine, \\
  Nanyang Technological University\\ 
  {\tt\small $\dagger$ Co-first authors, \tt\small $*$ Corresponding authors}
}

\begin{document}

\maketitle

\begin{abstract}
    Vision-and-language models (VLMs) have been increasingly explored in the medical domain, particularly following the success of CLIP in general domain. However, unlike the relatively straightforward pairing of 2D images and text, curating large-scale paired data in the medical field for volumetric modalities such as CT scans remains a challenging and time-intensive process. This difficulty often limits the performance on downstream tasks. To address these challenges, we propose a novel vision-language pre-training (VLP) framework, termed as \textbf{VELVET-Med}, specifically designed for limited volumetric data such as 3D CT and associated radiology reports. Instead of relying on large-scale data collection, our method focuses on the development of effective pre-training objectives and model architectures. The key contributions are: 1) We incorporate uni-modal self-supervised learning into VLP framework, which are often underexplored in the existing literature. 2) We propose a novel language encoder, termed as \textbf{TriBERT}, for learning multi-level textual semantics. 3) We devise the hierarchical contrastive learning to capture multi-level vision-language correspondence. Using only 38,875 scan-report pairs, our approach seeks to uncover rich spatial and semantic relationships embedded in volumetric medical images and corresponding clinical narratives, thereby enhancing the generalization ability of the learned encoders. The resulting encoders exhibit strong transferability, achieving state-of-the-art performance across a wide range of downstream tasks, including 3D segmentation, cross-modal retrieval, visual question answering, and report generation.
\end{abstract}

\section{Introduction}
\label{sec:intro}
The challenge of data scarcity is particularly pronounced in the medical domain, where concerns over patient privacy and the need for domain-specific annotation expertise further complicate data acquisition. To address this issue, a growing number of studies have adopted self-supervised learning approaches to pre-train domain-specific feature extractors, aiming to improve the accuracy and robustness of clinical tasks in the absence of large annotated datasets \cite{huang2021gloria, wang2022medclip, zhang2025medunifier, chen2023towards}. Building on the success of CLIP \cite{radford2021learning}, which effectively bridges visual and textual modalities, pre-training strategies on vision-and-language models (VLMs) has shown significant promise for learning generalizable encoders applicable to a variety of downstream tasks, such as image classification \cite{zhou2022conditional, zhou2022learning, huang2021gloria, wang2022medclip} and segmentation \cite{huang2021gloria, xu2022groupvit}. However, existing research has predominantly focused on two-dimensional imaging modalities, such as chest X-rays and histopathology slides \cite{huang2021gloria, bannur2023learning, javed2024cplip, Lu_2023_CVPR}, largely due to the availability of corresponding multi-modal datasets \cite{johnson2019mimic, ikezogwo2023quilt}. In contrast, the adaptation of such models to volumetric imaging data (e.g., CT and MRI scans) remains relatively underexplored. A key reason for this gap lies in the greater complexity of curating and managing volume–text pairs, which are more diagnostically nuanced and data-intensive than image–text counterparts.

Recent efforts have been directed toward constructing scan–report paired datasets \cite{xie2023towards, bai2024m3d}; however, the scale of these datasets remains significantly smaller than that of image–text counterparts. Volumetric data presents more complex and densely packed visual information, which naturally necessitates a larger number of training samples for effective modelling. Despite the challenge posed by limited parings, we argue that each volumetric sample inherently encapsulates richer and more comprehensive information than a 2D image. This insight underscores the importance of fully exploiting the volumetric nature of the data to learn meaningful representations. Rather than relying solely on large-scale datasets, this perspective motivates us to focus on designing robust visual pre-training strategies that can exploit the dense spatial context of 3D scans, thereby mitigating the impact of limited data availability.

Moreover, the main challenges in this field can be broadly categorized into two aspects. First, aligning global visual and textual features using contrastive loss or image-text matching loss \cite{li2021align} facilitates high-level and semantic understanding across data pairs. However, this approach often overlooks coarse, local information — an issue particularly pronounced in complex medical volumes and detailed diagnostic reports. Second, another significant challenge arises from the complexity and length of medical reports, which are considerably more elaborate than natural image captions. Unlike natural image descriptions — often limited to a few concise sentences — medical reports typically consist of multiple segments, each containing distinct clinical concepts that are interwoven and context-dependent, especially for complex volume-based report. Most existing approaches \cite{huang2021gloria, zhang2025medunifier} treat the entire report as a single sequence, extracting aggregated linguistic features using BERT-style language encoders \cite{devlin2019bert}, and thus overlooks the nuanced inter-sentence dependencies and the hierarchical structure that characterize medical narratives.



\begin{figure*}[tp]
    \centering{\includegraphics[width=\linewidth]{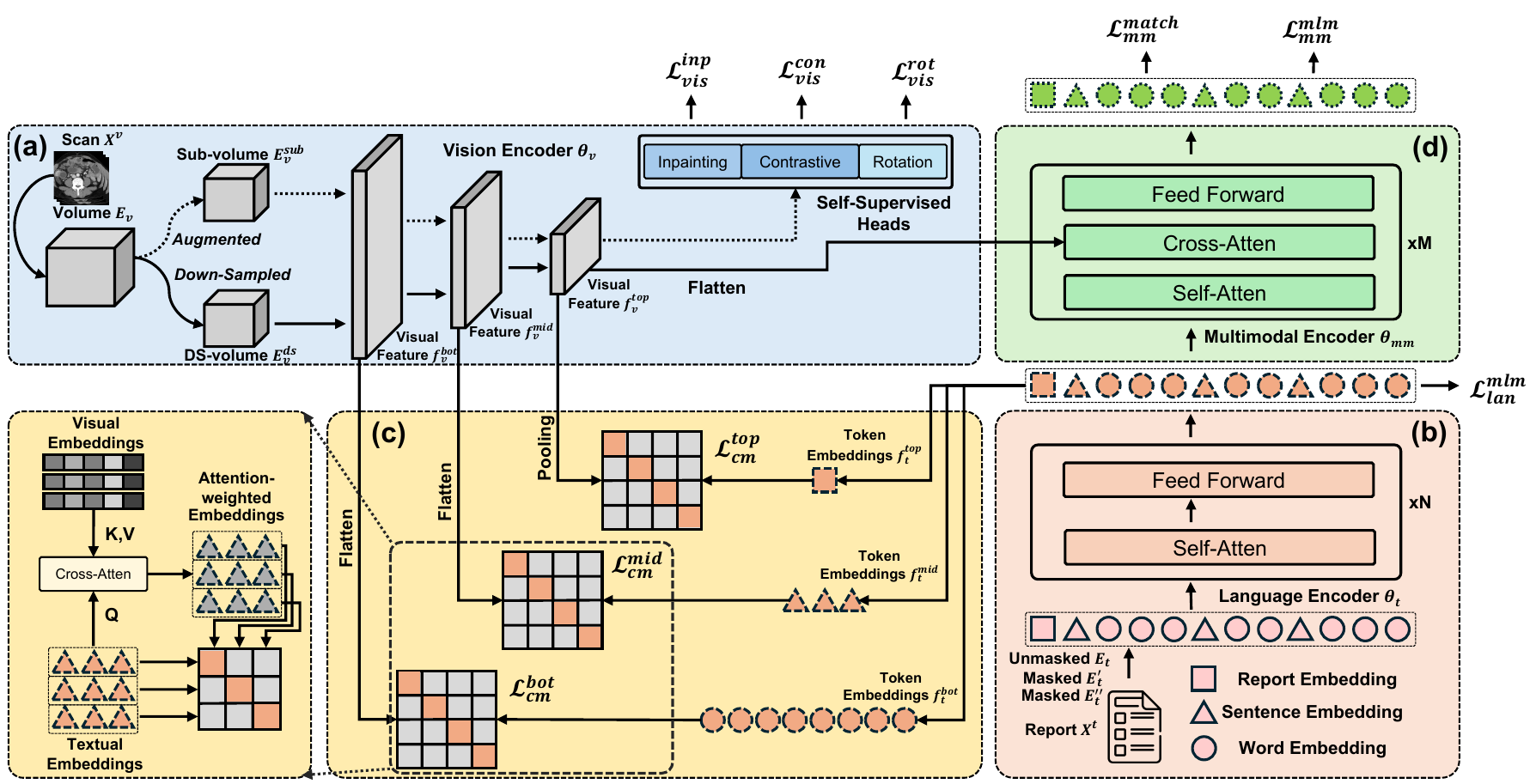}}
    \caption{Framework of VELVET-Med. The model takes paired CT scans and medical reports as input, leveraging four types of self-supervised learning signals: (a) uni-modal vision, (b) uni-modal language, (c) cross-modal, and (d) multi-modal. Detailed instruction can be found in \S\ref{subsec:objectives}. The input pair $(X^v, X^t)$ is transformed into tensors $E_v$ (Appendix \S\ref{sec:data_pre}) and $E_t$ (\S\ref{subsec:tribert}) for forward passing. 
    }
    \label{fig1}
\end{figure*}

To address the aforementioned challenges, we introduce \textbf{VELVET-Med}: \textbf{V}ision and \textbf{E}fficient \textbf{L}anguage pre-training for \textbf{V}olum\textbf{E}tric imaging \textbf{T}asks in \textbf{Med}icine, a lightweight and data-efficient framework tailored for volumetric medical scans. While the experiments in this article are conducted on 3D CT scans, the proposed framework is modality-agnostic and readily adatabple to other volumetric imaging data with corresponding diagnostic reports. The framework (see Figure \ref{fig1}) comprises a vision encoder, a language encoder, and a multi-modal encoder. These components, once trained, can function independently as feature extractors or be integrated as backbones for a variety of downstream tasks. With its hierarchical contrastive learning, novel TriBERT architecture, and supplementary uni-modal supervision signals, VELVET-Med effectively captures comprehensive vision-language features, achieving SOTA performance on medical vision-language benchmarks. The main contributions of this work are summarized as follows:



\begin{itemize}
    \item We introduce a novel and data-efficient VLP framework, VELVET-Med, tailored for medical volumetric data such as 3D CT scans and associated diagnostic reports. The resulting encoders, trained through this framework, exhibit significant performance gains over existing approaches across various downstream tasks and modalities.
    \item We propose a novel hierarchical contrastive learning method designed to align semantics across multiple granularities. This approach captures complex medical concepts associated with specific visual regions, thereby enhancing the accuracy of concept identification.
    \item We enhance the learning capacity of the BERT by introducing specialized input embeddings and an attention mask mechanism for sentence-level modelling. Our proposed Tri-Level BERT (TriBERT) facilitates the joint learning of report/sentence/word-level semantics. With our hierarchical contrastive learning, this design effectively bridges local and global representations, enabling a more comprehensive understanding of radiology reports.
    \item We integrate uni-modal self-supervision into the model to capture abstract cross-modal relationships while preserving detailed, fine-grained features within each individual modality.
    \item We also release a standardized multi-modal 3D CT dataset, M3D-CAP-filtered, to support reproducible evaluation of shared latent semantics learned by medical VLMs. This benchmark comprises carefully curated scans with high-quality, unambiguous volumetric data.

\end{itemize}

\section{Related work}
\label{sec:relared}

\subsection{Uni-modal self-supervised learning}
In the visual domain, self-supervised learning (SSL) has seen rapid advancement, driven by contrastive learning methods such as MoCo \cite{he2020momentum}, SimCLR \cite{chen2020simple}, and BYOL \cite{grill2020bootstrap}, as well as reconstruction-based approaches like image-MAE \cite{he2022masked} and iGPT \cite{chen2020generative}. These methods leverage large-scale, unlabeled image datasets to learn robust visual representations. Early SSL approaches relied on predefined proxy tasks, including solving jigsaw puzzles \cite{noroozi2016unsupervised}, colourization \cite{larsson2017colorization}, and rotation prediction \cite{feng2019self}. Over time, these evolved into more sophisticated techniques such as instance discrimination \cite{zhao2020makes} and masked auto-encoding \cite{he2022masked, reed2023scale}. These advances enable models to capture semantic information without the need for extensive labelled data, often matching or surpassing fully supervised baselines when fine-tuned on downstream tasks. From the perspective of the language domain, mainstream SSL methods can be broadly categorized into two paradigms. The first is masked language modelling, as introduced by BERT \cite{devlin2019bert}, which is primarily used for understanding tasks. The second is causal language modelling, derived from the original Transformer architecture \cite{vaswani2017attention}, and later scaled to large generative models such as GPT \cite{brown2020language} and LLaMA \cite{touvron2023llama}. In the medical domain, SSL has shown substantial potential for reducing dependence on labelled data, thereby facilitating model development in resource-constrained environments. Prominent studies \cite{chen2019med3d, zhou2021models, chen2023towards, zhang2025medunifier} have adapted SSL techniques to medical imaging modalities such as X-rays, as well as to medical text corpora. This growing body of work highlights the promise of SSL as a foundational approach for building scalable and generalizable models for tasks such as medical image analysis and clinical question answering.

\subsection{Vision-and-language model}
Vision-and-language models (VLMs) have advanced rapidly following the emergence of large-scale transformer architectures capable of jointly encoding image and text information. Early models such as ViLBERT \cite{lu2019vilbert}, LXMERT \cite{tan2019lxmert}, and UNITER \cite{chen2020uniter} introduced dual-stream or single-stream architectures to align visual and textual modalities, typically trained on large-scale web datasets using masked language modeling and image-text matching objectives. Building on this foundation, subsequent approaches — including OSCAR \cite{li2020oscar}, VinVL \cite{zhang2021vinvl}, CoCa \cite{yu2022cocacontrastivecaptionersimagetext}, CLIP \cite{radford2021learning}, and BLIP \cite{li2022blip, li2023blip} — have significantly pushed the state of the art by leveraging larger and more diverse datasets, enhanced contrastive learning strategies, and advanced multi-modal fusion techniques. Given the remarkable performance of VLMs in general-domain tasks, there has been growing interest in adapting these models to specialized domains, particularly in healthcare. In the medical domain, frameworks such as MedViLL \cite{moon2022multi} and BioViL \cite{bannur2023learning} extend general-purpose VLMs to integrate medical texts (e.g., radiology reports) with domain-specific imaging modalities (e.g., X-rays, CT scans, and MRIs). These adaptations involve the incorporation of domain knowledge and task-specific modules, enabling more accurate interpretation of complex medical data. The release of several medical image-text paired datasets — most notably MIMIC-CXR \cite{johnson2019mimic}, along with QUILT \cite{ikezogwo2023quilt} and PMC datasets \cite{zhang2023pmc} — has spurred the development of numerous models targeting 2D medical images and their associated reports \cite{wang2022medclip, zhang2025medunifier, chen2023towards, grill2020bootstrap}. However, 3D volumetric-text pairs remain underexplored, largely due to the high cost of data curation and the challenges associated with modelling complex spatial structures in three dimensions. In this work, we propose a novel model architecture and efficient self-supervised learning strategies tailored to volumetric CT scans and their corresponding medical reports, aiming to learn robust representations and enhance data efficiency for 3D medical imaging tasks. Additionally, while research works in the general domain has begun to explore the joint optimization of cross-modal and uni-modal self-supervised learning objectives \cite{mu2022slip, li2022supervisionexistseverywheredata}, this line of inquiry remains underexplored in the medical context. To address this gap, we integrate both uni-modal and cross-modal objectives and systematically evaluate their impact across a range of downstream tasks.

\section{Method}
\label{method}
In this section, we present our framework for the efficient pre-training of encoders on the CT scan modality. We begin by detailing the architecture used in the VLP in Section \S\ref{subsec:arch}, followed by a description of the diverse input employed for self-supervised tasks in Section \S\ref{subsec:prepare}. Next, we outline the self-supervised learning objectives in Section \S\ref{subsec:objectives}. Finally, we introduce the innovative TriBERT and the proposed hierarchical contrastive learning approach in Sections \S\ref{subsec:tribert} and \S\ref{subsec:hcl}, respectively.

\subsection{Model architecture}
\label{subsec:arch}
Inspired by \cite{li2021align}, our vision-and-language framework comprises three core encoders: a vision encoder $\theta_v$, a language encoder $\theta_t$, and a multi-modal encoder $\theta_{mm}$. Each mini-batch input consists of CT scans $X^{v} = \left\{x^v_{1}, x^v_{2}, \dots ,x^v_{N}\right\}$ paired with diagnostic notes $X^{t} = \left\{x^t_{1}, x^t_{2}, \dots ,x^t_{N}\right\}$. The $\theta_v$ processes $X^v$ to produce visual feature maps of shape $\mathbb{R}^{N \times c_v \times h \times w \times d}$, while the $\theta_t$ receives tokenized $X^t$ as sequences of embeddings in $\mathbb{R}^{N \times len \times c_t}$ (see \S\ref{subsec:tribert}). Here, $N$ is the batch size; $c_v$, $c_t$ are the feature dimensions; and $h \times w \times d$ denotes the spatial resolution of visual features. We adopt Swin-Transformer \cite{liu2021swin} for $\theta_v$ and the TriBERT (see \S\ref{subsec:tribert}) for $\theta_t$. The $\theta_{mm}$, implemented as transformer decoder layers stacked atop $\theta_t$, directly consumes its output without re-embedding.


\subsection{Preparation of model input}
\label{subsec:prepare}
\begin{figure*}[tp]
    \centering{\includegraphics[width=0.9\linewidth]{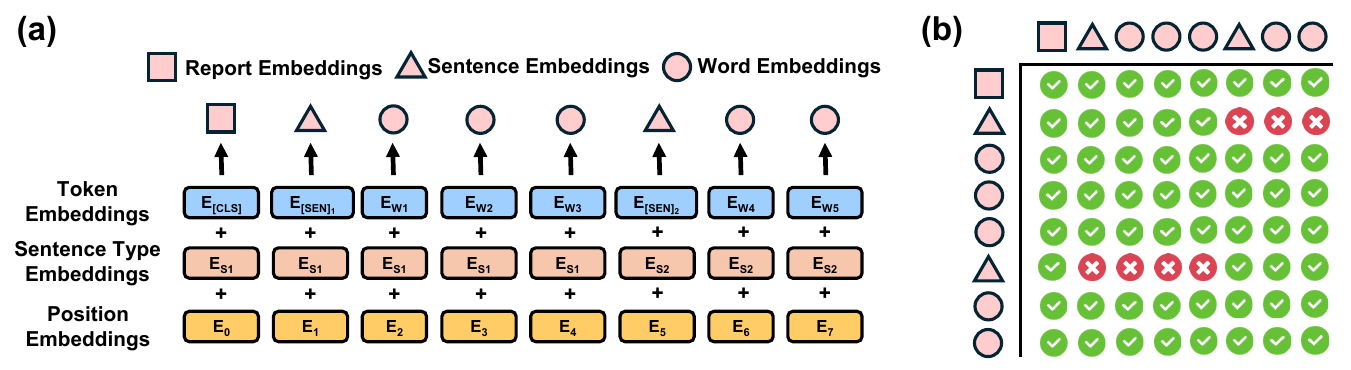}}
    \caption{Novel TriBERT design. \textbf{(a)} Input Embedding Construction: Unlike standard BERT, TriBERT inserts learnable [SENT$_{i}$] tokens into the input sequence to encode sentence semantics. It also adds sentence type embeddings to differentiate between sentences. \textbf{(b)} Tri-level Self-Attention Masking: To prevent inter-sentence information leakage, each [SENT$_{i}$] token attends only to its associated word tokens and the [CLS] token. This masking enforces sentence-level independence while retaining access to the global context.
    }
    \label{fig2}
\end{figure*}

Self-supervised tasks, guided by the inherent characteristics of each modality, require distinct input representations. For visual input, we downsample the full CT scan $E_v \in \mathbb{R}^{C \times H^{\prime} \times W^{\prime} \times D^{\prime}}$ to $E_v^{ds} \in \mathbb{R}^{C \times H \times W \times D}$ for cross- and multi-modal tasks, preserving overall anatomical structure. In contrast, vision self-supervised tasks use aggressive augmentations \cite{tang2022self} to extract a sub-volume $E_v^{sub} \in \mathbb{R}^{C \times H \times W \times D}$, capturing fine-grained local features. While $E_v^{ds}$ retains global context with reduced detail, $E_v^{sub}$ focuses on subtle anatomical variations at full resolution. This diversity in scale is essential for training a robust vision encoder. For textual input, we construct an embedding sequence $E_t \in \mathbb{R}^{N \times len \times c_t}$ (see \S\ref{subsec:tribert} and Figure \ref{fig2}). The unaltered $E_t$ is used in cross-modal tasks and multi-modal matching. Following the standard masking strategy from \cite{devlin2019bert}, we generate two masked variants, $E_t^{\prime}$ and $E_t^{\prime \prime}$, for uni-modal and multi-modal masked language modelling, respectively. 


\subsection{Learning objectives}
\label{subsec:objectives}
We introduce a set of learning objectives for optimization during the proposed VLP stage, categorized into four groups based on their specific goals and functions.

\paragraph{Uni-modal vision supervision.} Before the emergence of VLMs, numerous studies \cite{tang2022self, chen2022maskedimagemodelingadvances, 10658544} explored self-supervised learning on volumetric imaging to reduce the need for extensive manual labeling. These works demonstrated that vision-based self-supervised learning can yield effective feature extractors, enabling 3D clinical segmentation with limited annotated data. In this study, we hypothesize that combining VLP with vision self-supervised learning enhances model generalization, particularly under data-scarce conditions — a claim supported by strong empirical results in \S\ref{subsec:res}. Specifically, we employ masked volume inpainting \cite{DBLP:journals/corr/PathakKDDE16}, 3D rotation prediction \cite{DBLP:journals/corr/abs-1803-07728}, and contrastive coding \cite{DBLP:journals/corr/abs-1807-03748} as visual proxy tasks, resulting in the following objective:
\begin{align}
    \mathcal{L}_{vis} = \mathcal{L}_{vis}^{inp} +  \mathcal{L}_{vis}^{rot} + \mathcal{L}_{vis}^{con}
\end{align}

\paragraph{Uni-modal language supervision.} To better capture the complexity of clinical notes, we pre-train the language encoder $\theta_t$ using masked language modeling (MLM), relying solely on textual information. The objective function is defined as:
\begin{align}
    \mathcal{L}_{lan} = \mathcal{L}_{lan}^{mlm}
\end{align}
We hypothesize that uni-modal language supervision prior to multi-modal training enhances $\theta_t$'s ability to model text, thereby improving downstream cross-modal learning (empirical results in \S\ref{subsec:res}). Note that the masked tokens used in uni-modal training differ from those in multi-modal supervision.

\paragraph{Cross-modal supervision.} It plays a central role in VLMs, aligning visual and linguistic inputs within a shared embedding space and bridging the gap between distinct modalities. To capture deep correlations between complex volumetric CT scans and intricate clinical notes, we extend conventional global cross-modal supervision — typically applied only to final-layer features — into a hierarchical supervision framework. Specifically, top-, middle-, and bottom-level visual feature maps from $\theta_v$ are aligned with report-, sentence-, and word-level token embeddings from $\theta_t$, respectively. The cross-modal self-supervised objective is defined as:
\begin{align}
    \mathcal{L}_{cm} = \mathcal{L}_{cm}^{top} + \mathcal{L}_{cm}^{mid} + \mathcal{L}_{cm}^{bot} \label{equ:1} 
\end{align}
The detailed derivation and analysis are provided in \S\ref{subsec:hcl}.

\paragraph{Multi-modal supervision.} It aims to integrate visual and textual features more profoundly, enhancing inter-modality interaction. Following the approach in \cite{li2021align}, we adopt the "fuse after alignment" principle, using a multi-modal encoder to combine aligned visual and linguistic information into fused representations. The [CLS] token, which aggregates information from all tokens, is employed to compute the multi-modal matching loss $\mathcal{L}_{mm}^{match}$ through hard sample mining within the mini-batch \cite{li2021align, li2022blip}. This objective helps the model distinguish between matched and unmatched pairs using binary classification. Additionally, by replacing certain word tokens with a learnable [MASK] token and predicting the original words, we apply the masked language modeling objective $\mathcal{L}_{mm}^{mlm}$ to improve the model's multi-modal understanding. The overall multi-modal loss is given by:
\begin{align}
    \mathcal{L}_{mm} = \mathcal{L}_{mm}^{match} +\mathcal{L}_{mm}^{mlm}
\end{align}

We provide a summary of all the learning objectives and present the ultimate loss function: 
\begin{align}
    \mathcal{L} &= \mathcal{L}_{vis} + \mathcal{L}_{lan} + \mathcal{L}_{cm} + \mathcal{L}_{mm} \label{equ:5}
\end{align}

\subsection{TriBERT}
\label{subsec:tribert}
Unlike natural image captions, diagnostic reports in medical imaging consist of multiple interconnected sentences, each conveying distinct meanings. Conventional language encoders, such as BERT, primarily capture semantics at the word and whole-report levels, overlooking sentence-level implications. To address this, we propose Tri-level BERT, abbreviated as \textbf{TriBERT}, a model that simultaneously learns multi-level textual semantics from clinical notes.

\paragraph{Formation of input embeddings.} First, the report is segmented into individual sentences based on punctuation and the removal of redundant empty lines. Each word is then tokenized using WordPiece \cite{DBLP:journals/corr/abs-2012-15524}. A learnable token, [SENT$_i$], is prepended to each sentence, where $i$ denotes the sentence index, and a [CLS] token is placed at the beginning to represent the entire report. Next, we introduce a sentence-type embedding layer of size $\text{max\_num}_{sent} \times c_t$, where the $i$-th embedding is added to all tokens in the $i$-th sentence to emphasize sentence-level distinctions. Finally, positional embeddings are added, and the resulting embeddings are passed into the transformer blocks (Figure \ref{fig2} (a)).

\begin{figure*}[!]
    \centering{\includegraphics[width=1.0\linewidth]{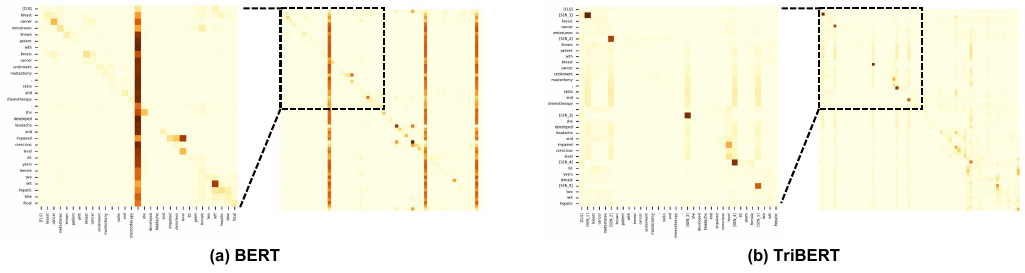}}
    \caption{Visualization of self-attention maps: self‑attention maps extracted from the final layer of the baseline BERT (left) and our TriBERT model (right); darker coloring denotes higher attention weights. The figure shows that BERT partially has meaningless tokens (e.g. a period) be highly attended over the context of medical report. Our novel TriBERT wipes out this unrelated learning pattern and reinforce intra-sentence correlations.}
    \label{fig3}
\end{figure*}

\paragraph{Tri-level self-attention mask.} The standard bi-directional self-attention mask introduces ambiguity in learning sentence-specific embeddings, as sentence tokens may attend to words outside their own sentence. To address this, we propose a novel self-attention mask: report and word-level tokens attend to the full context, while sentence-level tokens attend only to tokens within the same sentence and the global report token (Figure \ref{fig2} (b)). We present the visualization of self-attention maps for BERT and TriBERT in Figure \ref{fig3}. We found that BERT frequently assigns disproportionately high attention weights to punctuation tokens, such as periods, indicating spurious focus on uninformative semantics. The TriBERT alleviates this issue by attenuating attention to non‑semantic tokens and enhancing the modeling of meaningful intra‑sentence dependencies.

\subsection{Hierarchical contrastive learning}
\label{subsec:hcl}
The $\theta_v$ extracts a hierarchy of visual feature maps to align with token embeddings from the $\theta_t$. Specifically, $\theta_v$ outputs the last three feature maps, denoted as $f_v^k \in \mathbb{R}^{N \times c_v^k \times h^k \times w^k \times d^i}$ for $k \in \{top, mid, bot\}$. The top-level map $f_v^{\text{top}}$ is spatially average-pooled to obtain the final visual representation $f_v^{{top}} \in \mathbb{R}^{N \times c_v^{top}}$. On the language side, the [CLS] token embedding $f_t^{{rep}} \in \mathbb{R}^{N \times c_t}$ captures report-level semantics. Sentence-level semantics are represented by [SENT] token embeddings $f_t^{{sent}} \in \mathbb{R}^{N \times n_{{sent}} \times c_t}$, while word-level semantics $f_t^{{word}} \in \mathbb{R}^{N \times n_{{word}} \times c_t}$ are obtained by averaging sub-word token embeddings. For example, the word "pneumonia" may be split into "pneu" and "\#\#monia", which are aggregated to recover the full word representation \cite{DBLP:journals/corr/abs-2012-15524}. Finally, all representations $f$ are projected into a shared embedding space via linear transformation, yielding their aligned counterparts $g$. 

To align top-level semantics, the contrastive loss $\mathcal{L}_{cm}^{top}$ is computed between $g_v^{top}$ and $g_t^{rep}$ using the CLIP objective. For middle-level alignment, contextualized sentence embeddings $g_{tc}^{sent}$ are generated via cross-attention, where $g_v^{mid}$ serves as the context (key and value) and $g_t^{sent}$ as the query \cite{huang2021gloria}. A bi-directional contrastive loss $\mathcal{L}_{cm}^{mid}$ is then applied between $g_{tc}^{sent}$ and $g_t^{sent}$. Similarly, the bottom-level contrastive loss $\mathcal{L}_{cm}^{bot}$ is calculated between $g_{tc}^{word}$ and $g_t^{word}$.

Contrastive learning enables dual neural models to capture low-frequency, abstract features, demonstrating strong performance in understanding \cite{radford2021learning, mu2022slip} and generative tasks \cite{chen2020generative, pmlr-v139-ramesh21a}. However, in medical volumetric imaging, high-frequency visual cues — such as spatial and anatomical details — are equally crucial and may not be adequately captured by a single CLIP objective. Our proposed hierarchical contrastive learning strategy addresses this limitation by enabling the vision and language encoders to dynamically extract features at multiple levels of granularity, from coarse medical concepts (e.g., organs or structures) to fine-grained patterns (e.g., tissues or trabeculae), thereby achieving broader semantic alignment than conventional approaches.


\section{Experiments}
\label{sec:exp}

\subsection{Pre-training dataset}

We utilize the large-scale multi-modal 3D CT dataset, M3D-CAP \cite{bai2024m3d}, to pre-train our vision-language framework. The official release contains 120,092 scan-report pairs. Upon inspection, we identify two main issues: (1) some scans have disordered, non-anatomical slice arrangements along the z-axis; (2) others contain too few slices, making resampling visually misleading. To address these, we first exclude disordered scans, yielding M3D-CAP-clean with 69,686 pairs. We then filter out scans with fewer than 48 slices, resulting in M3D-CAP-filtered, containing 38,875 pairs. We use this final dataset for pre-training process. Detailed preprocessing steps and analyses are provided in Appendix \S\ref{sec:data_pre}. We argue that disorganized scans fall outside the distribution of real-world medical data and may hinder model learning. Experiments in \S\ref{subsec:abl} also confirm this hypothesis. For sentence modelling in TriBERT, we split each report into sentences using punctuation and line breaks, discarding those with fewer than two words. Report statistics are summarized in Table \ref{table:report_stats}.

\subsection{Downstream tasks and evaluation metrics}

\paragraph{3D semantic segmentation.} The segmentation task plays a critical role in 3D medical image analysis due to its ability to accurately recognize and localize anatomical structures. We adopt SwinUNETR \cite{hatamizadeh2021swin} as our segmentation model, initializing its vision encoder with pre-trained weights. Experiments are conducted on public datasets: AbdomenCT-1K \cite{9497733} with 1000 annotated CT scans and CT-Organ \cite{rister2020ct} with 140 annotated CT scans, using the official splits. Performance is evaluated using the Dice similarity coefficient (Dice), 95th percentile Hausdorff Distance (HD95), and Surface Dice (also known as Normalized Surface Distance, NSD) \cite{nikolov2021deeplearningachieveclinically}.


\paragraph{Cross-modal retrieval.} In cross-modal retrieval, the model matches relevant CT scans and reports based on cosine similarity, typically involving two tasks: scan-to-report retrieval (SRR) and report-to-scan retrieval (RSR). Evaluation is conducted on 2000 high-quality test set using recall at ranks 1, 5, and 10 to measure the model’s ability to retrieve relevant scans or reports.

\paragraph{Visual question answering.} Visual Question Answering (VQA) tasks are typically categorized into generative and classification-based VQA. Generative VQA requires producing textual answers based on visual input and accompanying questions. Performance is assessed using BLEU \cite{papineni-etal-2002-bleu}, ROUGE \cite{ROUGE}, and BERT-Score \cite{zhang2019bertscore}. In this work, we use the M3D-VQA dataset \cite{bai2024m3d} and implement generative VQA by combining a large language model (LLM) as a text decoder with our pre-trained VLM framework. Questions and associated volumetric data are encoded by the VLM to obtain prompt embeddings (i.e., outputs of the multi-modal encoder). These embeddings are projected via a linear transformation to match the LLM’s embedding dimension and are prepended to the decoder input to guide answer generation. During fine-tuning multi-modal encoder, linear projection as well as LLM are updated while freezing vision/language encoders. For classification-based VQA, where answers are selected from a fixed set (e.g., yes/no), we use the yes/no subset of M3D-VQA. A MLP classifier is added on top of the multi-modal encoder, using the [CLS] token for prediction. Evaluation is conducted using standard classification metrics, including AUC and Accuracy.

\paragraph{Medical report generation.} For report generation, the model takes a 3D medical vision data and a series of descriptive commands as input. Entire descriptive commands are listed in Appendix \S\ref{sec:instruction}. The VLM encodes both into prompt embeddings, which are then decoded by the LLM to generate the medical report. We fine-tune the model using the M3D-CAP-filtered dataset and evaluate it on the same test set used in the cross-modal retrieval task. During fine-tuning multi-modal encoder, linear projection as well as LLM are updated while freezing vision/language encoders. The evaluation metrics are consistent with those used in the generative VQA task.

\subsection{Implementation details}
For pre-training, we adopt the Swin-Transformer \cite{liu2021swin} as the vision encoder, the TriBERT as the language encoder, and a stack of transformer decoder blocks as the multi-modal encoder for our VLM. During pre-training, the sizes of $E_v^{ds}$ and $E_v^{sub}$ are set to $\mathbb{R}^{1 \times 96 \times 96 \times 96}$. Text inputs are constrained to a maximum of 50 sentences, 200 words per sentence, and 512 words per report. Pre-training is conducted for 50 epochs using the AdamW optimizer \cite{loshchilov2019decoupledweightdecayregularization} with cosine annealing, an initial learning rate of 2$e$-5, $\beta_1=0.9$, $\beta_2=0.98$, weight decay of 1$e$-5, and a batch of 10 per GPU. Experiments are run on 4 NVIDIA H100 GPUs. Each pre-training experiment takes 50 hours. For all downstream tasks, we employ the AdamW optimizer \cite{loshchilov2019decoupledweightdecayregularization} with a cosine annealing scheduler. In 3D semantic segmentation, the initial learning rate is set to 4$e$-4 for 300k iterations. For cross-modal retrieval, we evaluate models by ranking dot-product similarity between scans and reports on the held-out test set. In generative VQA and report generation, we use a 7B medicine-domain LLM \cite{ContactDoctor_Bio-Medical-Llama-3-8B} as the language decoder. We adopt LORA \cite{hu2022lora} for parameter-efficient fine-tuning with a rank of 8, a $\alpha$ of 32. The learning rate is set to 1$e$-4 for 10 epochs. For the classification-based VQA, we use an learning rate of 1$e$-4 for 5 epochs. Detailed implementation can be found in Appendix \S\ref{subsec:detailed_preimp}, \ref{subsec:detailed_dtimp}.

\subsection{Ablation study}
\label{subsec:abl}

\paragraph{Vision/language encoder configuration.} Encoder size, or the number of parameters, significantly affects model capacity. To identify the optimal size, we train vision and language encoders using top-level contrastive learning across varying encoder configurations (see Appendix \S\ref{subsec:enc_config}). SwinVIT-B has less parameter counts than the standard VIT-B by 27M and TriBERT-B has comparable parameter counts to the standard BERT-B used in the M3D-CAP models \cite{bai2024m3d}. Evaluation on the scan-report retrieval task (Table \ref{table:encoder_size}) shows that the combination of TriBERT-B and SwinVIT-S achieves the best performance while containing less trainable parameters and compute demands (FLOPs) compared to a legendary combination of BERT-B and VIT-B.


\begin{table}[!hbt]
\centering
\caption{Ablation study on encoder configurations. We assess cross-modal retrieval performance across various encoder configurations. The combination of TriBERT-B, SwinVIT-S yields the best results while including less parameters and FLOPs than the classical combination (BERT-B, VIT-B).}
\label{table:encoder_size}
\resizebox{0.95\textwidth}{!}{%
\begin{tabular}{@{}l|ccc|cccccc@{}}
\toprule
\multirow{2}{*}{Encoder configurations} & \multirow{2}{*}{\begin{tabular}[c]{@{}c@{}}scan size\\ (H,W,D)\end{tabular}} & \multirow{2}{*}{\#params} & \multirow{2}{*}{FLOPs} & \multicolumn{3}{c}{Scan-Report Retrieval (SRR)}  & \multicolumn{3}{c}{Report-Scan Retrieval (RSR)}  \\ \cmidrule(l){5-10} 
                                        &                                                                              &                           &                        & R@1            & R@5            & R@10           & R@1            & R@5            & R@10           \\ \midrule
BERT-B, VIT-B                           & 256,256,32                                                                   & 109M+115M                 & 439.56G                & 26.29          & 57.46          & 69.91          & 25.87          & 57.36          & 70.22          \\
TriBERT-B, VIT-B                        & 256,256,32                                                                   & 109M+115M                 & 442.86G                & 27.59          & 56.29          & 70.01          & 27.73          & 58.65          & 70.79          \\
TriBERT-S, SwinVIT-T                    & 96,96,96                                                                     & 70M+8M                    & 98.29G                 & 12.91          & 33.64          & 47.17          & 12.45          & 32.36          & 45.58          \\
TriBERT-B, SwinVIT-T                    & 96,96,96                                                                     & 109+8M                    & 146.62G                & 28.76          & 56.07          & 67.75          & 25.93          & 54.22          & 66.51          \\
TriBERT-B, SwinVIT-S                    & 96,96,96                                                                     & 109+22M                   & 206.30G                & \textbf{32.77} & \textbf{62.09} & \textbf{73.51} & \textbf{33.18} & \textbf{59.93} & \textbf{71.86} \\
TriBERT-B, SwinVIT-B                    & 96,96,96                                                                     & 109+88M                   & 444.75G                & 28.60          & 59.57          & 72.33          & 27.83          & 58.44          & 71.66          \\ \bottomrule
\end{tabular}%
}
\end{table}

\paragraph{Effect of data quality.} To also evaluate the impact of data quality, we implement the CLIP-3D model following the original paper \cite{bai2024m3d} and train it on the M3D-CAP-filtered dataset. As shown in Table \ref{table:retrieval}, our CLIP-3D model significantly outperforms the M3D retrieval model trained on the original M3D-CAP data, demonstrating the substantial benefits of using high-quality, unambiguous volumetric data. Both models share the same architecture — BERT-B for language encoding and VIT-B for vision encoding — differing only in their pre-training datasets.


\subsection{Results of downstream tasks}
\label{subsec:res}
To evaluate the effectiveness of VELVET-Med, we conduct experiments on V-L benchmark comprising 3D semantic segmentation, cross-modal retrieval, VQA, and report generation. We pre-train the framework using various combinations of self-supervised objectives and test the resulting models across tasks to analyze the impact of each objective. We begin with top-level cross-modal supervision ($\mathcal{L}_{cm}^{top}$), analogous to the CLIP paradigm \cite{radford2021learning}, and incrementally incorporate hierarchical cross-modal and uni-modal supervision. For cross/multi-modal unified supervision, we start with top-level cross-modal and multi-modal signals ($\mathcal{L}_{cm}^{top}, \mathcal{L}_{mm}$), following \cite{li2021align, chen2023towards}, and progressively add hierarchical cross-modal and uni-modal objectives, culminating in the VELVET-Med ($\mathcal{L}_{cm}, \mathcal{L}_{mm}, \mathcal{L}_{lan}, \mathcal{L}_{vis}$).


\paragraph{3D semantic segmentation.} We classify segmentation models into four categories: 1) Uni-modal baselines rely solely on visual data, either training the entire segmentor from scratch or using a vision-supervised pre-trained encoder. 2) Interactive segmentation, similar to SAM-style models \cite{kirillov2023segment, du2024segvol, ma2024segment}, incorporates complex prompt encoders based on geometry and language. 3) Cross-modal VLMs exclude multi-modal encoder and supervision, relying instead on pre-training with cross- or uni-modal supervision. 4) 
Cross/multi-modal VLMs are initialized with $\mathcal{L}_{cm}^{top}$ and $\mathcal{L}_{mm}$, then progressively incorporate uni-modal supervision and hierarchical contrastive learning. All models in categories 3) and 4) use SwinUNETR-S with different vision encoder weights via pre-training schemes. In Tables \ref{table:seg_res}, we report segmentation results on the AbdomenCT-1k and CT-ORG datasets. First, initializing the vision encoder from VELVET-Med yields the best performance, highlighting the effectiveness of our framework for 3D semantic segmentation. Second, emerging interactive segmentation models show performance drops — 15.0\% and 7.3\% Dice on AbdomenCT-1k and CT-ORG, respectively — compared to the best results, indicating that interactive segmentation still has a room for improvement in medical volumetric tasks. Complete results can be seen in Table \ref{table:seg_abd} and \ref{table:seg_ctorg} in Appendix \S\ref{subsec:add_seg}.



\begin{table}[!hbt]
\centering
\caption{Results on AbdomenCT-1K and CT-ORG using Dice(\%), NSD(\%) and HD95 for evaluating model's performance. The best/second-best results are highlighted in bold/underlined, respectively.}
\label{table:seg_res}
\resizebox{0.85\textwidth}{!}{%
\begin{tabular}{@{}clccc|ccc@{}}
\toprule
\multicolumn{2}{c}{\multirow{2}{*}{Methods}}                                                                         & \multicolumn{3}{c|}{AbdomenCT-1K} & \multicolumn{3}{c}{CT-ORG} \\
\multicolumn{2}{c}{}                                                                                                 & Dice(\%)    & NSD(\%)   & HD95    & Dice(\%) & NSD(\%) & HD95  \\ \midrule
\multirow{3}{*}{\begin{tabular}[c]{@{}c@{}}Uni-modal\\ baseline\end{tabular}}       & 
SwinUNETR-T (from scratch) \cite{hatamizadeh2021swin}     & 89.66       & 89.46     & 18.44   & 84.31    & 72.85   & 57.63 \\  & 
SwinUNETR-S (from scratch)     & 93.91       & 95.93     & 3.07    & 83.80    & 72.29   & 61.31 \\  & 
SwinUNETR-T (pre-trained by $\mathcal{L}_{vis}$) \cite{tang2022self} & \textbf{94.06}       & \underline{96.15}     & 3.07    & 84.75    & 74.95   & 58.05 \\ \midrule
\multirow{3}{*}{\begin{tabular}[c]{@{}c@{}}Interactive\\ segmentators\end{tabular}} & 
SegVol  \cite{du2024segvol}                         & 79.06       & -         & -       & 77.78    & -       & -     \\  & 
M3D segmentator (Linear) \cite{bai2024m3d}                   & 73.64       & -         & -       & 81.27    & -       & -     \\ & 
M3D segmentator (MLP) \cite{bai2024m3d}                      & 73.37       & -         & -       & 81.10    & -       & -     \\ \midrule
\multirow{5}{*}{\begin{tabular}[c]{@{}c@{}}Cross-modal\\ VLMs\end{tabular}}          & 
$\mathcal{L}_{cm}^{top}$  (CLIP-like) \cite{radford2021learning}                     & 93.92       & 95.89     & 2.93    & 86.09    & 77.24   & 42.71 \\  & 
$\mathcal{L}_{cm}$                          & 93.85       & 95.94     & 3.59    & 85.04    & 74.52   & 55.40 \\  & 
$\mathcal{L}_{cm},\mathcal{L}_{lan}$                  & 93.95       & 95.92     & 3.38    & 82.30    & 70.16   & 64.89 \\  & 
$\mathcal{L}_{cm},\mathcal{L}_{vis}$                   & 93.96       & 96.03     & \underline{2.71}    & 86.70    & 77.83   & 42.98 \\  & 
$\mathcal{L}_{cm},\mathcal{L}_{lan}, \mathcal{L}_{vis}$           & 93.22       & 94.86     & 4.28    & 84.47    & 72.61   & 56.78 \\ \midrule
\multirow{3}{*}{\begin{tabular}[c]{@{}c@{}}Cross/multi-modal\\ VLMs\end{tabular}}    & 
$\mathcal{L}_{cm}^{top}, \mathcal{L}_{mm}$              & 93.20       & 94.85     & 4.90    & \underline{87.33}   & \underline{79.66}   & 43.96 \\  & 
$\mathcal{L}_{cm}, \mathcal{L}_{mm}$                  & 93.95       & 95.98     & 4.01    & 86.33    & 77.35   & \underline{41.78} \\  & 
$\mathcal{L}_{cm},\mathcal{L}_{mm},\mathcal{L}_{lan},\mathcal{L}_{vis}$ (VELVET-Med)   & \underline{94.03}       & \textbf{96.20}     & \textbf{2.39}    & \textbf{88.30}    & \textbf{82.09}   & \textbf{37.97} \\ \bottomrule
\end{tabular}%
}
\end{table}

\paragraph{Cross-modal retrieval.} We adopt three baseline models for the retrieval task: PMC-CLIP \cite{lin2023pmc} (2D model), the M3D retrieval model \cite{bai2024m3d} trained on the full M3D-CAP dataset, and our own implementation of \cite{bai2024m3d}, called as CLIP-3D, pre-trained on M3D-CAP-filtered. Our proposed VLM framework includes two variants — cross-modal VLMs and cross/multi-modal VLMs — both pre-trained on M3D-CAP-filtered to evaluate different pre-training objectives. As shown in Table \ref{table:retrieval}, several key observations emerge. First, CLIP-3D outperforms the M3D retrieval model, highlighting the importance of high-quality, unambiguous pre-training data. Second, our VLM trained with $\mathcal{L}_{cm}^{top}$ surpasses CLIP-3D, indicating that SwinVIT-S is a more effective visual encoder for volumetric CT scans than ViT-B, despite five times fewer parameters. Moreover, hierarchical contrastive learning yields a ~3\% improvement across metrics compared to top-level cross-modal supervision. Adding uni-modal supervision provides marginal gains over using only cross-modal supervision. Interestingly, using both $\mathcal{L}_{cm}^{top}$ and $\mathcal{L}_{mm}$ leads to worse performance than using $\mathcal{L}_{cm}^{top}$ alone, suggesting that multi-modal supervision may hinder cross-modal learning in the CT scans. Ultimately, the VELVET-Med, integrating hierarchical contrastive and uni-modal supervision, achieves competitive retrieval results.


\begin{table}[]
\centering
\caption{Cross-modal retrieval results on test set. The top K (1, 5, 10) Recall metrics are reported. }
\label{table:retrieval}
\resizebox{0.85\textwidth}{!}{%
\begin{tabular}{@{}clcccccc@{}}
\toprule
\multicolumn{2}{c}{\multirow{2}{*}{Methods}}                                                                    & \multicolumn{3}{c}{Scan-Report Retrieval (SRR)} & \multicolumn{3}{c}{Report-Scan Retrieval (RSR)} \\ \cmidrule(l){3-8} 
\multicolumn{2}{c}{}                                                                                            & R@1            & R@5            & R@10          & R@1            & R@5            & R@10          \\ \midrule
\multirow{3}{*}{Baselines}                                                       & 
PMC-CLIP \cite{lin2023pmc}                    & 1.15           & 4.35           & 7.60          & 3.15           & 8.55           & 13.55         \\ & 
M3D retrieval model \cite{bai2024m3d}                         & 19.10          & 47.45          & 62.65         & 18.45          & 47.30          & 62.15         \\ & 
CLIP-3D                      & 26.29          & 57.46          & 69.91         & 25.87          & 57.36          & 70.22         \\ \midrule
\multirow{7}{*}{\begin{tabular}[c]{@{}c@{}}Cross-modal\\ VLMs\end{tabular}}       & 
$\mathcal{L}_{cm}^{top}$                   & 32.77          & 62.09          & 73.51         & 33.18          & 59.93          & 71.86         \\ & 
$\mathcal{L}_{cm}^{top}, \mathcal{L}_{cm}^{mid}$              & 34.57          & 63.22          & 74.49         & 33.95          & 63.21          & 73.93         \\
 & 
$\mathcal{L}_{cm}^{top}, \mathcal{L}_{cm}^{bot}$               & 34.67          & 64.40          & 75.62         & 34.26          & 63.48          & 73.15        \\& 
$\mathcal{L}_{cm}$                        & 35.96          & 63.68          & 75.67         & 35.39          & 63.63          & 75.05         \\& 
$\mathcal{L}_{cm}, \mathcal{L}_{lan}$                & 35.24          & \underline{66.36}          & 75.93         & 35.85          & \underline{64.66}          & 74.95         \\& 
$\mathcal{L}_{cm}, \mathcal{L}_{vis}$                & \textbf{36.85}          & 65.85          & 74.79         & \underline{35.37}          & \textbf{64.95}          & 74.92         \\& 
$\mathcal{L}_{cm}, \mathcal{L}_{lan}, \mathcal{L}_{vis}$        & \underline{36.18}          & \textbf{66.54}         & \textbf{77.43}         & \textbf{36.60}          & 64.49          & \underline{75.07}        \\ \midrule
\multirow{3}{*}{\begin{tabular}[c]{@{}c@{}}Cross/multi-modal\\ VLMs\end{tabular}} & 
$\mathcal{L}_{cm}^{top}, \mathcal{L}_{mm}$            & 27.67          & 58.38          & 70.01         & 26.80          & 57.87          & 70.22         \\& 
$\mathcal{L}_{cm}, \mathcal{L}_{mm}$                 & 33.02          & 61.99          & 72.69         & 31.48          & 60.60          & 71.19         \\& 
$\mathcal{L}_{cm}, \mathcal{L}_{vis}, \mathcal{L}_{lan}, \mathcal{L}_{vis}$ (VELVET-Med)& 34.86          & 65.72          & \underline{76.66}         & 33.12          & 64.44          & \textbf{75.11}         \\ \bottomrule
\end{tabular}%
}
\end{table}

\begin{figure*}[!htb]
    \centering{\includegraphics[width=0.9\linewidth]{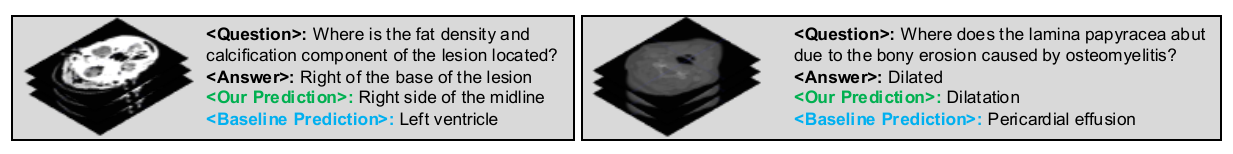}}
    \caption{Qualitative comparisons with VELVET-MED and ground truth on open-ended VQA.}
    \label{fig4}
\end{figure*}

\paragraph{Visual question answering.} For generative VQA, we re-implemented M3D-LaMed \cite{bai2024m3d} as a baseline for comparison with our proposed method, incorporating several adjustments for a fair evaluation. Specifically, we use a pre-trained vision encoder from CLIP-3D, followed by a 3D perceiver, to transform CT scans into a sequence of visual embeddings. These visual embeddings are then concatenated with language instruction embeddings to create unified prompt embeddings, which serve as inputs for generating medical reports or answers. As shown in Table \ref{table:vqa}, our VLM, trained with $\mathcal{L}_{cm}^{top}$, $\mathcal{L}_{mm}$, outperforms the baseline across most metrics. Adding hierarchical contrastive learning ($\mathcal{L}_{cm}$, $\mathcal{L}_{mm}$) provides further improvement, while the VELVET-Med model pre-trained with the complete objectives achieves the best overall performance. In Figure \ref{fig4}, we present the qualitative comparisons between VELVET-MED and baseline for generative, open-ended VQA, proving model's understanding and generative capabilities. More information can be seen in \S\ref{subsec:add_vqa}. The results of classification-based VQA show in Table \ref{table:vqa} where the VELVET-Med model achieves the best performance in terms of both AUC and accuracy.

\paragraph{Medical report generation.} We adopt the same models and schemes used in the generative VQA to evaluate report generation. Table \ref{table:vqa} indicates that our VLM, optimized with $\mathcal{L}_{cm}^{top}$ and $\mathcal{L}_{mm}$, surpasses the baseline on BLEU, BERT-Score. The full VELVET‑Med achieves the strongest results.

\begin{table}[]
\centering
\caption{Results on report generation and visual question answering. Generative tasks allow evaluation with BLEU, ROUGE, and BERT-Score. In contrast, classification-based VQA is framed as a fixed-set classification task, where AUC and accuracy are the primary metrics. $\$$ indicates our re-implementation of M3D-LaMed \cite{bai2024m3d} with the pre-trained vision encoder from CLIP-3D.}
\label{table:vqa}
\resizebox{0.85\textwidth}{!}{%
\begin{tabular}{@{}lcccccccc@{}}
\toprule
\multirow{2}{*}{Methods}     & \multicolumn{3}{c}{Report Generation} & \multicolumn{3}{c}{Generative VQA} & \multicolumn{2}{c}{Cls-based VQA} \\ \cmidrule(l){2-9} 
                             & BLEU      & ROUGE     & BERT-Score    & BLEU     & ROUGE    & BERT-Score   & AUC              & Acc             \\ \midrule
M3D-LaMed $^\$$                   & 17.32     & 11.73     & 84.09         & \underline{20.63}    & 23.47    & 87.40        & -                & -               \\ \midrule
$\mathcal{L}_{cm}^{top}, \mathcal{L}_{mm}$            & 18.46     & 11.47     & \underline{84.98}         & 18.93    & 29.67    & 88.32        & \underline{84.01}            & 75.94           \\
$\mathcal{L}_{cm}, \mathcal{L}_{mm}$                 & \underline{21.42}     & \underline{12.27}     & 84.90         & 20.49    & \underline{30.24}    & \underline{88.41}        & 83.85            & \underline{76.12}          \\
$\mathcal{L}_{cm}, \mathcal{L}_{vis}, \mathcal{L}_{lan}, \mathcal{L}_{vis}$ (VELVET-Med) & \textbf{23.54}     & \textbf{13.28}     & \textbf{85.16}         &    \textbf{22.93}     &    \textbf{34.75}      &       \textbf{88.96}      &     \textbf{84.27}             &     \textbf{76.17}            \\ \bottomrule
\end{tabular}%
}
\end{table}

\section{Conclusion}
In this paper, we propose VELVET-Med, a vision and efficient language pre-training for volumetric imaging tasks in medicine. By integrating TriBERT with hierarchical contrastive learning, our approach effectively aligns visual and textual semantics at multiple levels, leveraging the fine-grained structure of scan-report pairings. We also highlight the critical role of uni-modal supervision in learning robust, generalizable neural models. Our ablation studies reveal two key findings: (1) model size significantly impacts performance, especially in data-scarce scenarios, where larger models can sometimes degrade results; (2) high-quality data can substantially outperform larger, noisier datasets collected through web crawling. Finally, VELVET-Med sets a new benchmark for medical CT scan understanding across diverse vision-and-language tasks, paving the way for scaling VLM to broader, web-scale and well-curated medical datasets in the future.

\medskip

{
\small
\bibliographystyle{unsrt}
\bibliography{reference}
}

\appendix
\newpage
\appendix

\renewcommand{\thetable}{\Alph{table}}
\setcounter{table}{0} 

\renewcommand{\thefigure}{\Alph{figure}}
\setcounter{figure}{0} 

\section{3D CT scan pre-processing}
\label{sec:data_pre}

\label{subsec:3dct}
\begin{figure}[!htb]
    \centering{\includegraphics[width=\linewidth]{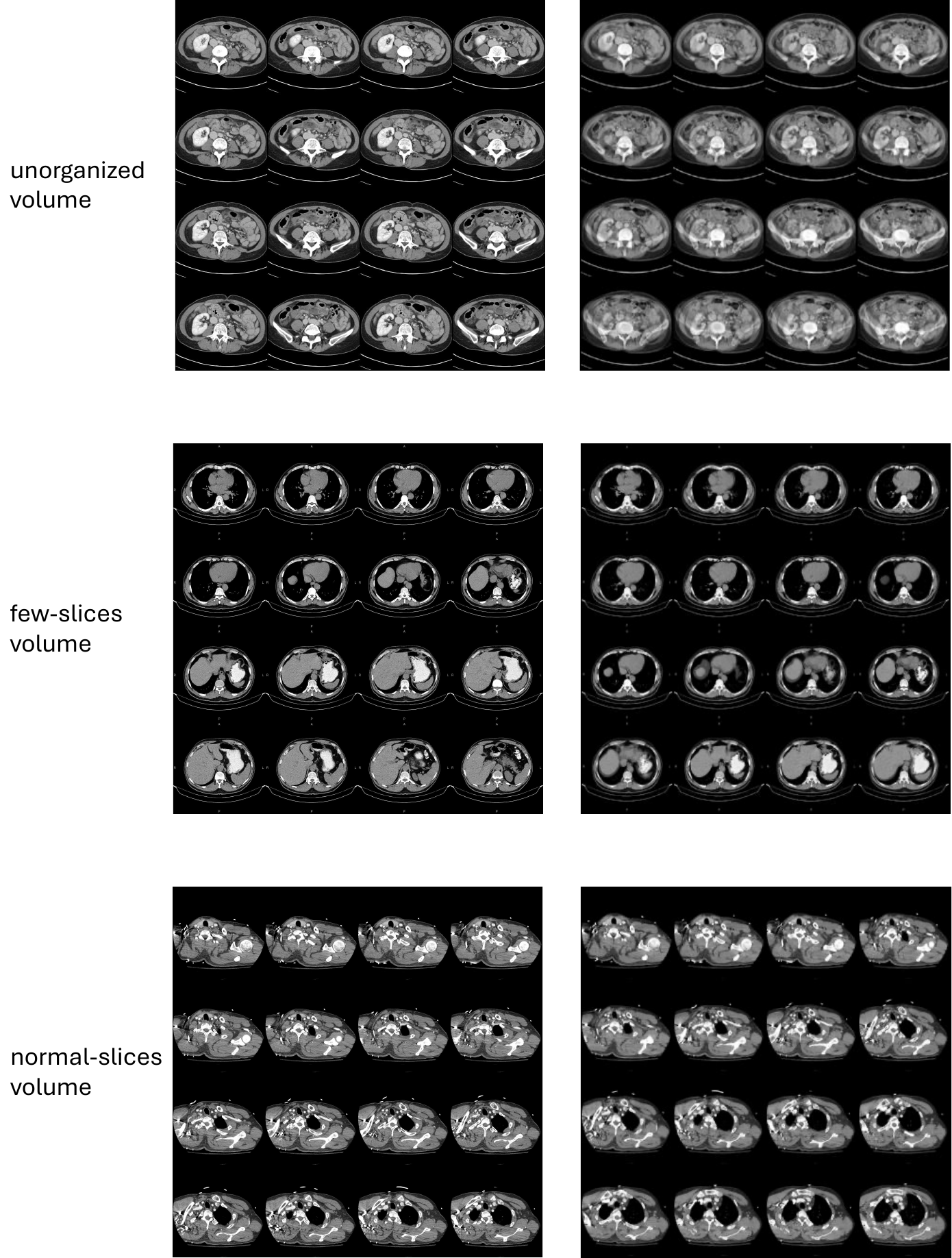}}
    \caption{Illustration of different CT scan types. The first column shows the original volumes, while the second displays the resampled versions used as input for the vision encoder. Each panel presents a sequence of slices arranged from the top-left to the bottom-right along the z-axis. In the top row, slices are unordered and non-anatomical, leading to a visually disorganized resampled volume (\textbf{color}). In the middle row, the original volume has fewer slices (e.g., 48) than required by the vision encoder (e.g., 96). Resampling to match the input size introduces slight distortions in some regions (\textbf{color}). In the bottom row, the original volume has more slices (e.g., 120) than needed. Downsampling at uniform intervals along the z-axis preserves anatomical consistency, avoiding any visual confusion.}        
    \label{fig:ill_volume}
\end{figure}

Upon inspecting the multi-modal CT scan dataset M3D-CAP \cite{bai2024m3d}, we observed that scans are stored as sequences of 2D slices along the z-axis without accompanying metadata, deviating from standard formats like DICOM or NIfTI. Many slice stacks lack anatomical order, resulting in disorganized volumes when simply stacked (see Figure \ref{fig:ill_volume}). Additionally, some scans contain fewer slices than typical CT volumes. Naïve interpolation along the z-axis introduces visual artifacts and ambiguity (see Figure \ref{fig:ill_volume}, top row), hindering feature learning and generalization of the vision encoder. To address this, we exclude unorganized scans from M3D-CAP, reducing the dataset from 120,092 to 69,686 pairings, which we term M3D-CAP-clean. Within M3D-CAP-clean, some scans still contain fewer than 10 slices, potentially impairing unimodal self-supervised learning due to repeated or zero-padded regions. To mitigate this, we retain only scans with more than 48 slices, resulting in the M3D-CAP-filtered dataset (38,875 pairings). Slice count statistics for both datasets are provided in Table \ref{table:resolution}. For scans with more than 96 slices, we uniformly sample 96 slices along the z-axis to form the full volume $E_v$. For scans with more than 48 while less than 96 slices, we upsample along the z-axis by a factor of 2 to obtain $E_v$. The volume $E_v$ is resized in the x- and y-dimensions using 2D interpolation [cite] and downsampled along the z-axis to generate the global volume $E_v^{ds}$. For vision self-supervised learning, we extract and augment sub-volumes $E_v^{sub}$ from $E_v$.

\begin{table}[!]
\centering

\caption{Statistics of medical reports from M3D-CAP-filtered. \# of words/sentences represent word counts and sentence counts within each report. Length of sentence is word counts of each sentence.}
\label{table:report_stats}

\resizebox{\textwidth}{!}{%
\begin{tabular}{@{}c|cccccc@{}}
\toprule
                          & max   & min & mean  & 25\% quartiles & median & 75\% quartiles \\ \midrule
\# of words               & 687.0 & 4.0 & 109.9 & 71.0           & 89.0   & 137.0          \\
\# of sentences           & 61.0  & 2.0 & 10.5  & 7.0            & 9.0    & 12.0           \\
max length of sentence    & 127.0 & 2.0 & 25.5  & 19.0           & 24.0   & 30.0           \\
min length of sentence    & 2.0   & 1.0 & 2.3   & 2.0            & 3.0    & 11.0           \\
mean length of sentence   & 44.0  & 2.0 & 10.5  & 8.5            & 10.3   & 12.1           \\
median length of sentence & 47.0  & 1.0 & 9.3   & 7.0            & 9.0    & 11.0           \\ \bottomrule
\end{tabular}%
}
\end{table}


\begin{table}[]
\centering
\caption{Statistics about the number of slices for M3D-CAP-clean and M3D-CAP-filtered. We exclude volume with less than 48 slices out of M3D-CAP-clean to form our pre-training dataset M3D-CAP-filtered, where each scan contains rich z-axis information. }
\label{table:resolution}

\resizebox{\textwidth}{!}{%
\begin{tabular}{@{}c|cccccc@{}}
\toprule
                 & max    & min  & mean  & 25\% quartiles & median & 75\% quartiles \\ \midrule
M3D-CAP-clean    & 1210.0 & 1.0  & 69.1  & 27.0           & 54.0   & 92.0           \\
M3D-CAP-filtered & 1210.0 & 48.0 & 106.9 & 64.0           & 86.0   & 119.0          \\ \bottomrule
\end{tabular}%
}
\end{table}

\section{Additional implementation details}
\label{sec:add_imple}

\subsection{Vision/language encoder configuration}
\label{subsec:enc_config}
A set of configuration and the number of parameters is as follows:
\begin{itemize}
    \item SwinVIT-T: params: 8M; embed\_dim: 48; depth: [2, 2, 2, 2]; num\_heads: [3, 6, 12, 24]
    \item SwinVIT-S: params: 22M; embed\_dim: 48; depth: [2, 8, 8, 8]; num\_heads: [4, 8, 16, 32]
    \item SwinVIT-B: params: 88M; embed\_dim: 96; depth: [2, 8, 8, 8]; num\_heads: [4, 8, 16, 32]
    \item TriBERT-S: params: 70M; feature\_dim: 768; num\_layers: 6
    \item TriBERT-B: params: 109M; feature\_dim: 768; num\_layers: 12
\end{itemize}
For SwinVIT embed\_dim represents input feature dimension for stage one, depth and num\_heads represent the number of Swin-Transformer blocks and attention heads for each stage respectively. For TriBERT feature\_dim represents feature dimension for all token embeddings, num\_layers is the number of transformer encoder layers. Note that TriBERT and BERT has similar model structure and comparable parameters with the key distinction of input embedding construction and attention mask mechanism.

\subsection{Pre-training implementations}
\label{subsec:detailed_preimp}
We adopt the Swin-Transformer \cite{liu2021swin} as the vision encoder, TriBERT as the text encoder, and a stack of transformer decoder blocks as the multi-modal encoder for our VLM. During pre-training, the sizes of $E_v^{ds}$ and $E_v^{sub}$ are set to $\mathbb{R}^{1 \times 96 \times 96 \times 96}$. Text inputs are constrained to a maximum of 50 sentences, 200 words per sentence, and 512 tokens per report. Pre-processing details for CT scans are provided in Appendix \S\ref{sec:data_pre}. For cross-modal supervision, we use a 512-dimensional embedding for multi-level joint spaces. A binary cross-entropy loss is applied for multi-modal matching, with hard samples mined via the top-level contrastive similarity matrix \cite{li2021align, li2022blip}. We adopt a 15\% masking ratio for uni- and multi-modal masked language modeling \cite{devlin2019bert, li2021align}. For uni-modal vision supervision, we implement three self-supervised objectives: (1) Masked volume inpainting with a 30\% sub-volume drop rate and a shallow upsampling stack for recovery; (2) 3D contrastive coding using a 512-dimensional embedding and InfoNCE loss \cite{DBLP:journals/corr/abs-1807-03748}; (3) Rotation prediction as a four-class classification task with rotation angles of $0^{\circ}$, $90^{\circ}$, $180^{\circ}$, and $270^{\circ}$.  Pre-training is conducted for 50 epochs using the AdamW optimizer \cite{loshchilov2019decoupledweightdecayregularization} with cosine annealing, an initial learning rate of 2$e$-5, $\beta_1=0.9$, $\beta_2=0.98$, weight decay of 1$e$-5, and a batch size of 10 per GPU. The best model is selected based on validation loss. Experiments are run on 4 NVIDIA H100 GPUs using Accelerate \cite{accelerate} with mixed precision and data parallelism.

\subsection{Implementation of downstream tasks}
\label{subsec:detailed_dtimp}
For all downstream tasks, we fine-tune using the AdamW optimizer \cite{loshchilov2019decoupledweightdecayregularization} with a cosine annealing scheduler. In 3D semantic segmentation, the initial learning rate is set to 4$e$-4 for 300k iterations, and sliding window inference is applied during testing to generate full-volume mask predictions. For cross-modal retrieval, we evaluate pre-trained models without fine-tuning by ranking dot-product similarity between CT scans and reports on the M3D-CAP-filtered test set. In generative VQA and medical report generation tasks, we use a 7B medicine-domain LLM \cite{ContactDoctor_Bio-Medical-Llama-3-8B} as the language decoder, with a linear projection aligning the feature dimensions between the VLM and LLM. We adopt LORA \cite{hu2022lora} for parameter-efficient fine-tuning with a rank of 8, a $\alpha$ of 32. The initial learning rate is set to 1$e$-4 for 10 epochs. For the classification-based VQA task, we use an initial learning rate of 1$e$-4, trained for 5 epochs. Due to time constraints and computational scarcity, we sample 10\% of M3D-VQA and M3D-CAP-filtered to fine-tune for open-ended VQA and report generation.

\section{Additional results}
\label{sec:add_res}

\subsection{3D semantic segmentation}
\label{subsec:add_seg}

To complement the concise summary reported in the main manuscript, Table \ref{table:seg_abd} and Table \ref{table:seg_ctorg} provide organ‑level evaluation of our model on the multi‑organ CT benchmark. Each table lists the mean Dice coefficient for every anatomical structure together with two surface–based metrics, Normalized Surface Dice (NSD) and the 95th‑percentile Hausdorff distance (HD95), offering a more granular view of boundary quality.

\begin{table}[t]
\centering
\caption{Results on AbdomenCT-1K. Dice(\%), NSD(\%) and HD95 are utilzied to evaluate model's performance. The best and second-
best results are highlighted in bold and underlined, respectively. Full De-MedViL achieves the best performance on NSD and HD95.}
\label{table:seg_abd}
\resizebox{\textwidth}{!}{%
\begin{tabular}{@{}clccccccc@{}}
\toprule
\multicolumn{2}{c}{\multirow{2}{*}{Methods}}                                                                         & \multicolumn{5}{c}{Dice (\%) $\uparrow$}               & \multirow{2}{*}{NSD (\%) $\uparrow$} & \multirow{2}{*}{HD95 $\downarrow$} \\ \cmidrule(lr){3-7}
\multicolumn{2}{c}{}                                                                                                 & Liver & Kidney & Spleen & Pancreas & Avg   &                          &                       \\ \midrule
\multirow{3}{*}{\begin{tabular}[c]{@{}c@{}}Uni-modal\\ baseline\end{tabular}}       & 
SwinUNETR-T (from scratch) \cite{hatamizadeh2021swin}    & 96.59 & 93.35  & 93.34  & 75.35    & 89.66 & 89.46                    & 18.44                 \\ & SwinUNETR-S (from scratch)     & 97.81 & 96.35  & 97.12  & 84.36    & 93.91 & 95.93                    & 3.07                  \\  & SwinUNETR-T (pre-trained by $\mathcal{L}_{vis}$) \cite{tang2022self} & 97.80 & 96.17  & \underline{97.17}  & \textbf{85.11}   & \textbf{94.06} & \underline{96.15}                    & 3.07                  \\ \midrule
\multirow{3}{*}{\begin{tabular}[c]{@{}c@{}}Interactive\\ segmentation\end{tabular}} & 
SegVol  \cite{du2024segvol}                &    -   &   -     &  -      &   -       & 79.06 &       -                   & -                      \\ 
& M3D segmentator (Linear) \cite{bai2024m3d}                  &  -     &     -   &     -   &   -       & 73.64 &         -                 &   -                    \\ & 
M3D segmentator (MLP)  \cite{bai2024m3d}                    &    -   &    -    &      -  &      -    & 73.37 &     -                     &    -                   \\ \midrule
\multirow{5}{*}{\begin{tabular}[c]{@{}c@{}}Cross-modal\\ VLMs\end{tabular}}          & 
$\mathcal{L}_{cm}^{top}$  (CLIP-like) \cite{radford2021learning}                  & 97.79 & 96.16  & 97.14  & 84.59    & 93.92 & 95.89                    & 2.93                  \\  & 
$\mathcal{L}_{cm}$                          & 97.77 & 95.94  & 97.12  & 84.55    & 93.85 & 95.94                    & 3.59                  \\ & 
$\mathcal{L}_{cm}, \mathcal{L}_{lan}$                  & \textbf{97.83} & 96.20  & 97.12  & 84.65    & 93.95 & 95.92                    & 3.38                  \\  & 
$\mathcal{L}_{cm}, \mathcal{L}_{vis}$                  & 97.78 & \underline{96.36}  & 97.17  & 84.53    & 93.96 & 96.03                    & \underline{2.71}                  \\  & 
$\mathcal{L}_{cm}, \mathcal{L}_{lan}, \mathcal{L}_{vis}$          & 97.59 & 96.07  & 96.88  & 82.35    & 93.22 & 94.86                    & 4.28                  \\ \midrule
\multirow{3}{*}{\begin{tabular}[c]{@{}c@{}}Cross/multi-modal\\ VLMs\end{tabular}}    & 
$\mathcal{L}_{cm}^{top},\mathcal{L}_{mm}$           & 97.56 & 95.50  & 96.89  & 82.84    & 93.20 & 94.85                    & 4.90                  \\   & 
$\mathcal{L}_{cm},\mathcal{L}_{mm}$                   & 97.81 & 96.23  & \textbf{97.18}  & 84.56    & 93.95 & 95.98                    & 4.01                  \\ & $\mathcal{L}_{cm},\mathcal{L}_{mm},\mathcal{L}_{lan},\mathcal{L}_{vis}$ (De-MedViL)  & \underline{97.81} & \textbf{96.38}  & 97.16  & \underline{84.80}    & \underline{94.03} & \textbf{96.20}                    & \textbf{2.39}                  \\ \bottomrule
\end{tabular}%
}
\end{table}

\begin{table}[t]
\centering
\caption{Results on CT-ORG. Dice(\%), NSD(\%) and HD95 are utilzied to evaluate model's performance. The best and second-
best results are highlighted in bold and underlined, respectively. Full De-MedViL achieves the best performance across varying metrics.}
\label{table:seg_ctorg}
\resizebox{\textwidth}{!}{%
\begin{tabular}{@{}clcccccccc@{}}
\toprule
\multicolumn{2}{c}{\multirow{2}{*}{Methods}}                                                                         & \multicolumn{6}{c}{Dice(\%) $\uparrow$}                      & \multirow{2}{*}{NSD(\%) $\uparrow$} & \multirow{2}{*}{HD95 $\downarrow$} \\ \cmidrule(lr){3-8}
\multicolumn{2}{c}{}                                                                                                 & Liver & Bladder & Lungs & Kidneys & Bone  & Avg   &                          &                       \\ \midrule
\multirow{3}{*}{\begin{tabular}[c]{@{}c@{}}Uni-modal\\ baselines\end{tabular}}      & 
SwinUNETR-T (from scratch) \cite{hatamizadeh2021swin}    & 90.08 & 71.40   & 94.89 & 80.34   & 86.79 & 84.31 & 72.85                    & 57.63                 \\  & 
SwinUNETR-S (from scratch)     & 87.93 & 71.42   & 94.98 & 80.39   & 86.20 & 83.80 & 72.29                    & 61.31                 \\ & 
SwinUNETR-T (pre-trained by $\mathcal{L}_{vis}$)\cite{tang2022self} & 89.17 & 73.54   & 95.08 & 80.43   & 87.39 & 84.75 & 74.95                    & 58.05                 \\ \midrule
\multirow{3}{*}{\begin{tabular}[c]{@{}c@{}}Interactive\\ segmentators\end{tabular}} & 
SegVol \cite{du2024segvol}                        & -     & -       & -     & -       & -     & 77.78 & -                        & -                     \\  & 
M3D segmentator (Linear) \cite{bai2024m3d}                   & -     & -       & -     & -       & -     & 81.27 & -                        & -                     \\  & 
M3D segmentator (MLP)  \cite{bai2024m3d}                    & -     & -       & -     & -       & -     & 81.10 & -                        & -                     \\ \midrule
\multirow{5}{*}{\begin{tabular}[c]{@{}c@{}}Cross-modal\\ VLMs\end{tabular}}          & 
$\mathcal{L}_{cm}^{top}$  (CLIP-like)\cite{radford2021learning}                 & 91.08 & 73.64   & 96.01 & 83.46   & 88.29 & 86.09 & 77.24                    & 42.71                 \\  & 
$\mathcal{L}_{cm}$                          & 89.49 & 74.03   & 95.66 & 80.03   & 87.99 & 85.04 & 74.52                    & 55.40                 \\  & 
$\mathcal{L}_{cm}, \mathcal{L}_{lan}$                 & 86.06 & 70.03   & 94.89 & 75.68   & 86.73 & 82.30 & 70.16                    & 64.89                 \\  & 
$\mathcal{L}_{cm}, \mathcal{L}_{vis}$                  & 91.65 & \underline{76.57}   & 95.68 & 83.08   & 88.67 & 86.70 & 77.83                    & 42.98                 \\  & 
$\mathcal{L}_{cm}, \mathcal{L}_{lan}, \mathcal{L}_{vis}$          & 89.07 & 73.07   & 95.42 & 80.48   & 86.32 & 84.47 & 72.61                    & 56.78                 \\ \midrule
\multirow{3}{*}{\begin{tabular}[c]{@{}c@{}}Cross/multi-modal\\ VLMs\end{tabular}}    & 
$\mathcal{L}_{cm}^{top}, \mathcal{L}_{mm}$               & \underline{93.02} & 76.33   & \underline{96.11} & \underline{84.60}   & \underline{88.67} & \underline{87.33} & \underline{79.66}                    & 43.96                 \\  & 
$\mathcal{L}_{cm}, \mathcal{L}_{mm}$                   & 90.58 & 76.26   & 96.02 & 82.45   & 88.42 & 86.33 & 77.35                    & \underline{41.78}                 \\  & 
$\mathcal{L}_{cm}, \mathcal{L}_{mm}, \mathcal{L}_{lan},\mathcal{L}_{vis}$ (De-MedViL)   & \textbf{93.24} & \textbf{78.40}   & \textbf{96.30} & \textbf{86.55}   & \textbf{89.23} & \textbf{88.30} & \textbf{82.09}                    & \textbf{37.97}                 \\ \bottomrule
\end{tabular}%
}
\end{table}

\subsection{Open-ended VQA}
\label{subsec:add_vqa}

\begin{figure*}[tp]
    \centering{\includegraphics[width=0.9\linewidth]{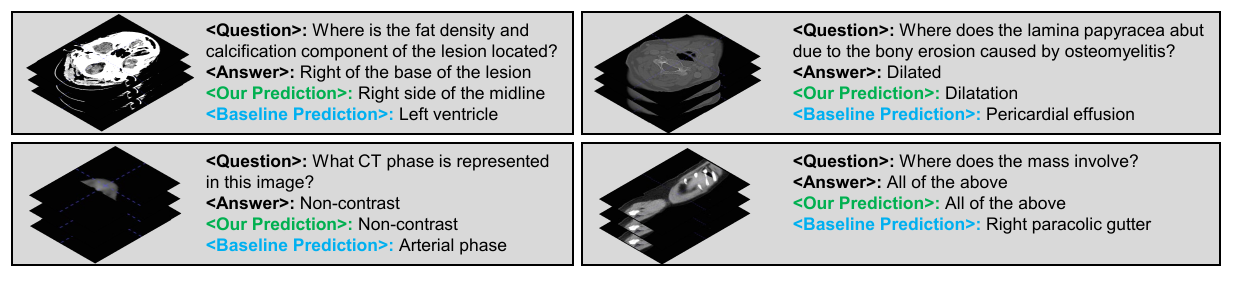}}
    \caption{Qualitative comparisons with VELVET-MED and ground truth on open-ended VQA.}
    \label{fig4_app}
\end{figure*}

In Figure \ref{fig4_app}, we provide qualitative comparison between the VELVET-MED and our implemented baseline model. It shows that VELVET-MED can produce more accurate answers.

\section{Instruction set for report generation}
\label{sec:instruction}
For fair comparison, we directly adopt instructions from \cite{bai2024m3d}. These instructions typically include prompts or guidelines for generating specific sections or content within the medical reports.
\begin{itemize}
    \item Can you provide a caption consists of findings for this medical image?
    \item  Describe the findings of the medical image you see.
    \item   Please caption this medical scan with findings.
    \item   What is the findings of this image?
    \item   Describe this medical scan with findings.
    \item   Please write a caption consists of findings for this image.
    \item   Can you summarize with findings the images presented?
    \item   Please caption this scan with findings.
    \item   Please provide a caption consists of findings for this medical image.
    \item   Can you provide a summary consists of findings of this radiograph?
    \item   What are the findings presented in this medical scan?
    \item   Please write a caption consists of findings for this scan.
    \item   Can you provide a description consists of findings of this medical scan?
    \item   Please caption this medical scan with findings.
    \item   Can you provide a caption consists of findings for this medical scan?
\end{itemize}

\section{Limitations}
\label{sec:lim}
Despite promising transferability to various downstream tasks, our study has several notable limitations that warrant consideration. First, pre-training was conducted on just 38k scan–report pairs, significantly smaller than the hundreds of millions of image–text pairs typically used in general-domain VLMs. This scale difference likely constrains the model’s ability to capture complex anatomical structures, varied linguistic expressions, and rare pathological patterns, contributing to its remaining performance gap relative to human experts on fine-grained clinical tasks. This underscores the need for resource-efficient VLP frameworks tailored for clinical applications. Additionally, VELVET-Med was developed using data from a single CT modality, potentially embedding domain-specific features that may limit generalization to other imaging types, such as MRI or ultrasound. Future work should prioritize assembling diverse, multi-institutional, multi-modal datasets, scaling pre-training to $10^7$ paired studies, incorporating self-distillation and curriculum learning to reduce annotation noise, and developing clinically grounded interpretability metrics.
  

\section{Broader impact}
\label{sec:impact}
Our 3D VLP framework offers significant advantages, including faster and more precise CT interpretation, reduced reporting burdens, improved access in resource-constrained settings, and a lower entry barrier for downstream medical-AI research. However, it also introduces notable risks: Dataset bias can undermine fairness across demographics and institutions, while over-reliance on automated outputs may promote clinical complacency or workforce displacement. Additionally, volumetric scans pose privacy risks by encoding identifiable anatomical features. Large-scale 3D pre-training is energy-intensive, and liability remains ambiguous when errors occur. Effective mitigation strategies include systematic bias audits on external cohorts, radiologist-in-the-loop deployment, privacy-preserving methods (e.g., federated or differentially private learning), energy-efficient training with transparent carbon reporting, and comprehensive model cards outlining intended use, limitations, and failure modes.


\end{document}